\documentclass[lettersize,journal]{IEEEtran}
\usepackage{amsmath,amsfonts}
\usepackage{algorithmic}
\usepackage{algorithm}
\usepackage{array}
\usepackage[caption=false,font=normalsize,labelfont=sf,textfont=sf]{subfig}
\usepackage{textcomp}
\usepackage{stfloats}
\usepackage{url}
\usepackage{verbatim}
\usepackage{graphicx}
\usepackage{cite}

\usepackage{subfloat}
\usepackage{booktabs}
\usepackage{multirow}
\usepackage{makecell}
\usepackage{lipsum}
\makeatletter
\let\NAT@parse\undefined
\makeatother
\usepackage{hyperref}  
\begin{document}

\title{Towards Semi-supervised Dual-modal\\ Semantic Segmentation}

\author{~Qiulei~Dong,~\IEEEmembership{Member,~IEEE,} ~Jianan~Li, ~Shuang~Deng
\thanks{Corresponding Author: Qiulei Dong.}
\thanks{Qiulei Dong and Jianan Li are with the School of Artificial Intelligence, University of Chinese Academy of Sciences, Beijing 100049, China, the State Key Laboratory of Multimodal Artificial Intelligence Systems, Institute of Automation, and also with the Center for Excellence in Brain Science and Intelligence Technology, Chinese Academy of Sciences, Beijing 100190, China (e-mail: qldong@nlpr.ia.ac.cn, lijianan211@mails.ucas.ac.cn).}
\thanks{Shuang Deng is with the Autonomous Driving Division of X Research Department, JD logistics, Beijing 102600, China (e-mail: dengshuang10@jd.com).}
}

\markboth{IEEE TRANSACTIONS ON Multimedia }%
{Shell \MakeLowercase{\textit{et al.}}: A Sample Article Using IEEEtran.cls for IEEE Journals}


\maketitle

\begin{abstract}
With the development of 3D and 2D data acquisition techniques, it has become easy to obtain point clouds and images of scenes simultaneously, which further facilitates dual-modal semantic segmentation. 
Most existing methods for simultaneously segmenting point clouds and images rely heavily on the quantity and quality of the labeled training data. 
However,  massive point-wise and pixel-wise labeling procedures are time-consuming and labor-intensive. 
To address this issue, we propose a parallel dual-stream network to handle the semi-supervised dual-modal semantic segmentation task, called PD-Net, by jointly utilizing a small number of labeled point clouds, a large number of unlabeled point clouds, and unlabeled images.
The proposed PD-Net consists of two parallel streams (called original stream and pseudo-label prediction stream). 
The pseudo-label prediction stream predicts the pseudo labels of unlabeled point clouds and their corresponding images.
Then, the unlabeled data is sent to the original stream for self-training. 
Each stream contains two encoder-decoder branches for 3D and 2D data respectively. 
In each stream, multiple dual-modal fusion modules are explored for fusing the dual-modal features.
In addition, a pseudo-label optimization module is explored to optimize the pseudo labels output by the pseudo-label prediction stream.
Experimental results on two public datasets demonstrate that the proposed PD-Net not only outperforms the comparative semi-supervised methods but also achieves competitive performances with some fully-supervised methods in most cases. 
\end{abstract}


\section{Introduction}
\begin{figure}
	\centering
	\includegraphics[width=1.00\columnwidth]{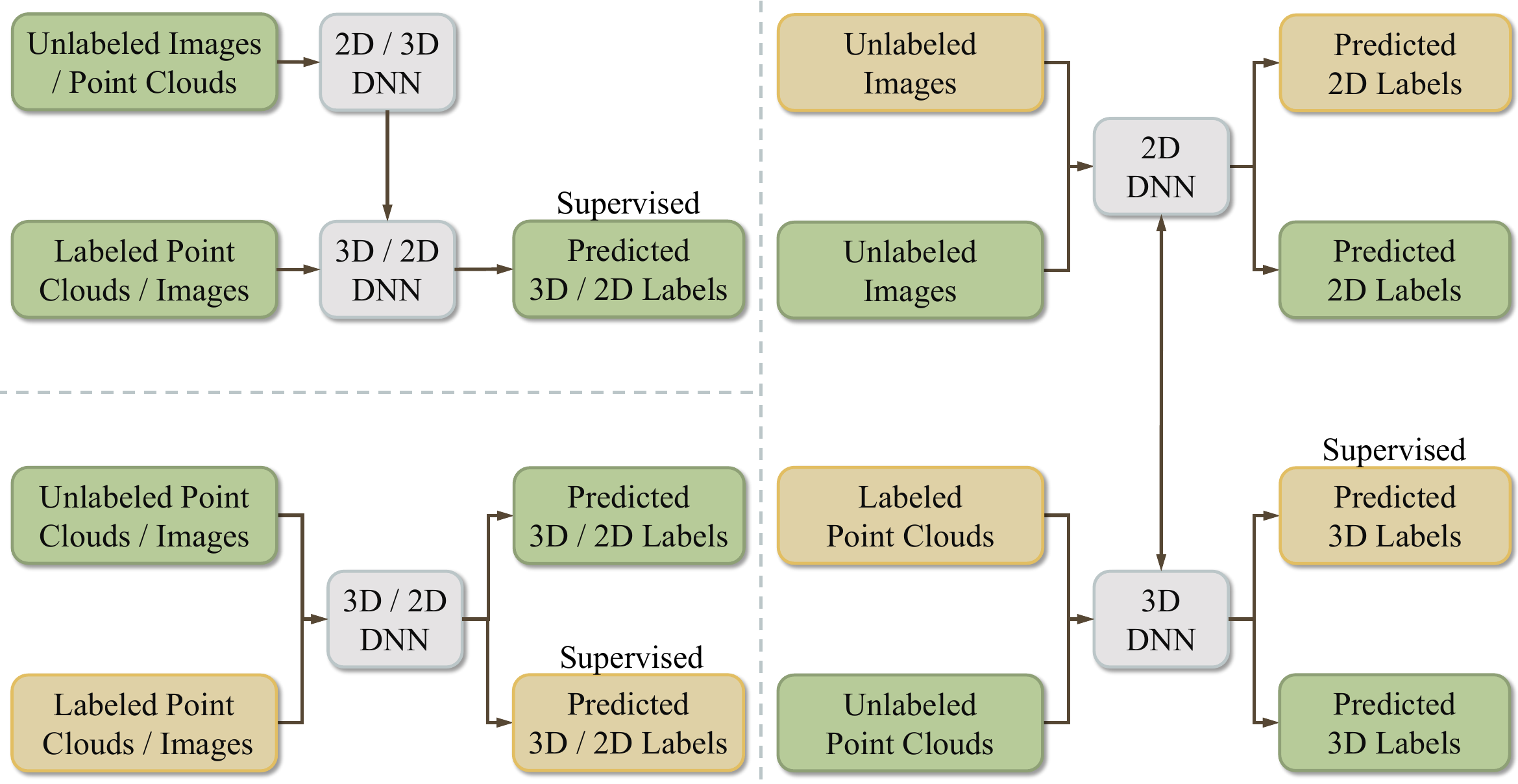}
	\caption{Diagrams of the fully-supervised dual-modal segmentation (top-left), semi-supervised uni-modal segmentation (bottom-left), and our semi-supervised dual-modal segmentation (right). In each framework, boxes in the same color represent they are corresponding with each other (i.e., point cloud with image / input with output). `Supervised' represents the output is supervised by the ground truth.}
	\label{task}
\end{figure}


\IEEEPARstart{W}{ith} the rapid development of both 3D and 2D data acquisition techniques, the 3D point clouds and images of scenes could be easily acquired together by jointly utilizing 3D and 2D sensors.
And the correspondences between 3D points and image pixels could be easily calculated with the intrinsic and extrinsic parameters of the sensors.
Accordingly, unlike the existing works \cite{9127813,2021ganet, 2021scfnet,10204183, 2021pointtransformer, tmm1, tmm2, tmm3} that only segment uni-modal data, many segmentation
methods \cite{20183dmv, 2019mvpnet, 2020fuseseg, Yan20222DPASS2P, Zhuang2021PerceptionAwareMF, zhao2023lif,10147273} are proposed to combine the complementary information of point clouds and images to boost performances, and they are trained in a fully-supervised manner under a general diagram shown in the top-left part of Figure~\ref{task}. 
However, these fully-supervised dual-modal segmentation methods generally require a time-consuming and labor-intensive labeling procedure.

In order to alleviate the above data annotation problem, some works \cite{2021guided, 2022sssnet, 2017pimodel, 2017meanteacher, 2020tcsmv2} focus on semi-supervised semantic segmentation for either 3D point clouds or 2D images, where a small proportion of the training data is labeled.
And the general diagram of these methods is shown in the bottom-left part of Figure~\ref{task}.
However, these semi-supervised methods only use uni-modal data, which could not make the most of the collected dual-modal data. 
Thus, how to utilize the complementary information in point clouds and images to solve the semi-supervised dual-modal segmentation problem remains to be investigated.

To address the aforementioned problems, we propose a parallel dual-stream network to simultaneously handle the semi-supervised semantic segmentation tasks for both point clouds and images, called PD-Net. 
It contains two parallel streams with the same architecture: an original stream, and a pseudo-label prediction stream whose parameters are updated by the Exponential Moving Average (EMA) strategy \cite{2017meanteacher}.
Each stream in PD-Net contains a 3D encoder-decoder branch, a 2D encoder-decoder branch, and multiple dual-modal fusion modules.
The 3D and 2D branches are utilized to extract 3D and 2D features respectively.
Intuitively, jointly leveraging dual-modal features could improve the segmentation performance, considering the complementarity of 3D features and 2D features (i.e., the 3D features contain rich geometric information but lack textural information, while the 2D features are enriched with color and textural information but are short of depth information).
However, direct fusion may dilute the inter-modal attentive weights, which could undermine the performance instead.
To fully exploit the complementary information in dual-modal data, we propose the dual-modal fusion module, which fuses the 3D and 2D latent features via a multi-head attention-based mechanism.
Besides, a consistency loss term is designed to constrain the semantic consistency between the 3D and 2D features.
The general diagram of the proposed PD-Net is illustrated in the right part of Figure~\ref{task},
it utilizes a small number of labeled point clouds, a large number of unlabeled point clouds, and unlabeled images for training. 
The labeled point clouds and their corresponding images are only trained in the original stream, and the labels of the images are projected from the point clouds according to the sensor parameters.
The unlabeled point clouds and their corresponding images are trained in both two streams.
Specifically, the output of the original stream is supervised by the pseudo labels output by the pseudo-label prediction stream.
To improve the quality of the pseudo labels generated by the pseudo-label prediction stream so that the effectiveness of the self-training strategy for the unlabeled point clouds and their corresponding images is guaranteed, we propose the pseudo-label optimization module to leverages pseudo labels of one modality to improve the quality of pseudo labels of another modality based on a voting mechanism. 
The pseudo-label optimization module is non-parametric, thus it is free from inductive bias and performance degeneration due to the domain gap between different modalities.


In sum, the main contributions of this paper include:
\begin{itemize}
\item We propose the dual-modal fusion module and the consistency loss term, which could effectively fuse the features of point clouds and images. 
\item We propose the pseudo-label optimization module, which is helpful for improving the quality of the predicted pseudo labels.
\item We propose the PD-Net, which consists of the aforementioned dual-modal fusion module, consistency loss term, and pseudo-label optimization module. To our best knowledge, this work is the first attempt to investigate how to utilize dual-modal data to handle the semi-supervised segmentation task for both point clouds and images.
\end{itemize}

The remainder of this paper is organized as follows. 
Some existing methods on 3D semi-supervised semantic segmentation, 2D semi-supervised semantic segmentation, and fully-supervised dual-modal semantic segmentation are reviewed in Section \ref{sec: rw}.
The proposed method is introduced in detail in Section \ref{sec: meth}.
The experimental results are reported in Section \ref{sec: exp}.
Finally, we conclude this paper in Section \ref{sec: con}.
\section{Related Works}
\label{sec: rw}
In this section, we first introduce the related semi-supervised segmentation methods of point clouds and images respectively.
Then, we introduce the related fully-supervised segmentation methods that combine the point clouds and images.

\subsection{3D Semi-supervised Semantic Segmentation}

To address the problem of semi-supervised semantic segmentation for 3D point clouds,
some early works \cite{2021redal,2021lepcss} rely on additional information (i.e., expert annotation) to constrain the features of unlabeled point clouds.
However, the application is limited because the introduced expert knowledge is not applicable to all circumstances.
To overcome this defect,
Li \textit{et al.}  \cite{2021spcs} proposed to design an adversarial architecture to calculate the confidence discrimination of pseudo labels for the unlabeled point clouds, and select the pseudo labels with high reliability.
Jiang \textit{et al.}  \cite{2021guided} proposed to utilize the contrastive loss based on the pseudo-label guidance to enhance the feature representation and model generalization ability in semi-supervised setting.
Deng \textit{et al.}  \cite{2022sssnet} proposed to combine the geometry and color-based superpoints to optimize the pseudo labels to guarantee the reliability of the self-training of the unlabeled points.
Taking the prior knowledge of LiDAR point clouds into consideration, Kong \textit{et al.} \cite{Kong_2023_CVPR} proposed to mix laser beams from different LiDAR scans and then encourage the model to make consistent and confident predictions before and after mixing.
Li \textit{et al.} \cite{LessIsMore} designed a soft pseudo-label method informed by LiDAR reflectivity to make full use of the limited labeled points and abundant unlabeled points.

\subsection{2D Semi-supervised Semantic Segmentation}

The great advances of semi-supervised learning in image classification \cite{2017pimodel, 2017meanteacher} inspire the investigation of semi-supervised semantic segmentation for images.
Early works on 2D semi-supervised semantic segmentation leveraged the Generative Adversarial Networks (GAN) to synthesize high-quality pseudo labels.
Hung \textit{et al.} \cite{Hung2018AdversarialLF} designed a fully-convolutional discriminator which enables semi-supervised learning by searching the reliable regions in predicted results of unlabeled images, thereby providing additional supervisory signals for training.
Mittal \textit{et al.} \cite{Mittal2019SemiSupervisedSS} proposed a GAN-based branch to improve the low-level details in segmentation predictions, which is helpful for alleviating low-level artifacts in the low-data regime.

Recently, researchers have paid more and more attention to consistency regularization and contrastive learning.
Chen \textit{et al.} \cite{Chen2021SemiSupervisedSS} imposed the consistency between networks with different initialization and encouraged the high similarity between the predictions of the two networks, which expands the training data by regarding the pseudo labels as the supervision for unlabeled images.
Liu \textit{et al.} \cite{Liu2021BootstrappingSS} proposed a contrastive learning framework designed at a regional level that performs semi-supervised pixel-level contrastive learning on a sparse set of hard negative pixels.
Alonso \textit{et al.} \cite{Alonso2021SemiSupervisedSS} maintained a memory bank that is updated across the whole dataset, and then enforced the network to yield similar pixel-level feature representations for same-class samples.
Wang \textit{et al.} \cite{Wang_2023_CVPR} proposed to apply regularization on the structure of the feature cluster, which is expected to increase the intra-class compactness in feature space.
Zhong \textit{et al.} \cite{Zhong2021PixelCS} combined consistency regularization and contrastive learning, which simultaneously constrains the label-space consistency property between images under different perturbations and the feature space contrastive property among different pixels.

\subsection{Fully-supervised Dual-modal Semantic Segmentation}

\begin{figure*}[!t]
	\centering
	\begin{tabular}{@{}c@{\extracolsep{0.1em}}c@{\extracolsep{0.1em}}c@{\extracolsep{0.1em}}c@{}}
		\includegraphics[width=0.6\linewidth]{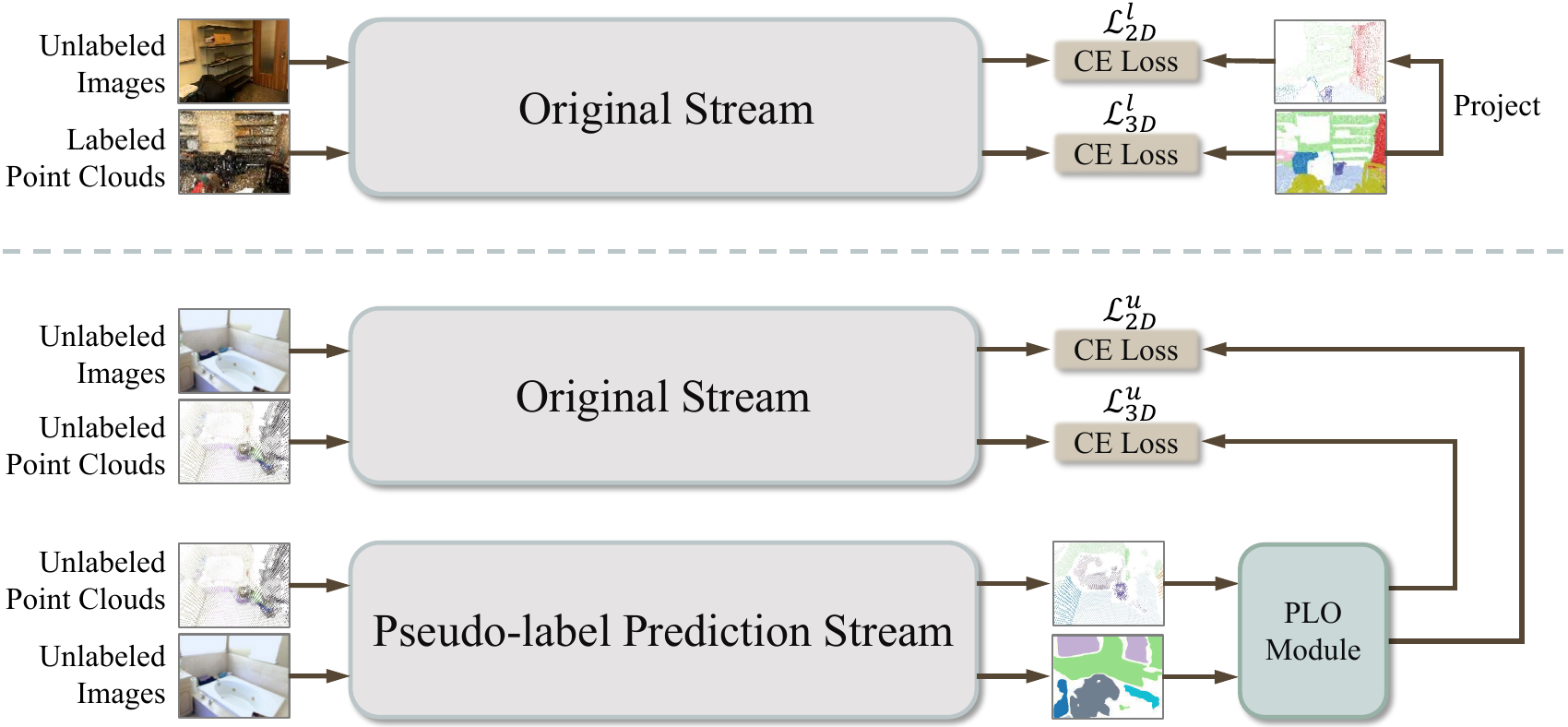}&
		\includegraphics[width=0.4\linewidth]{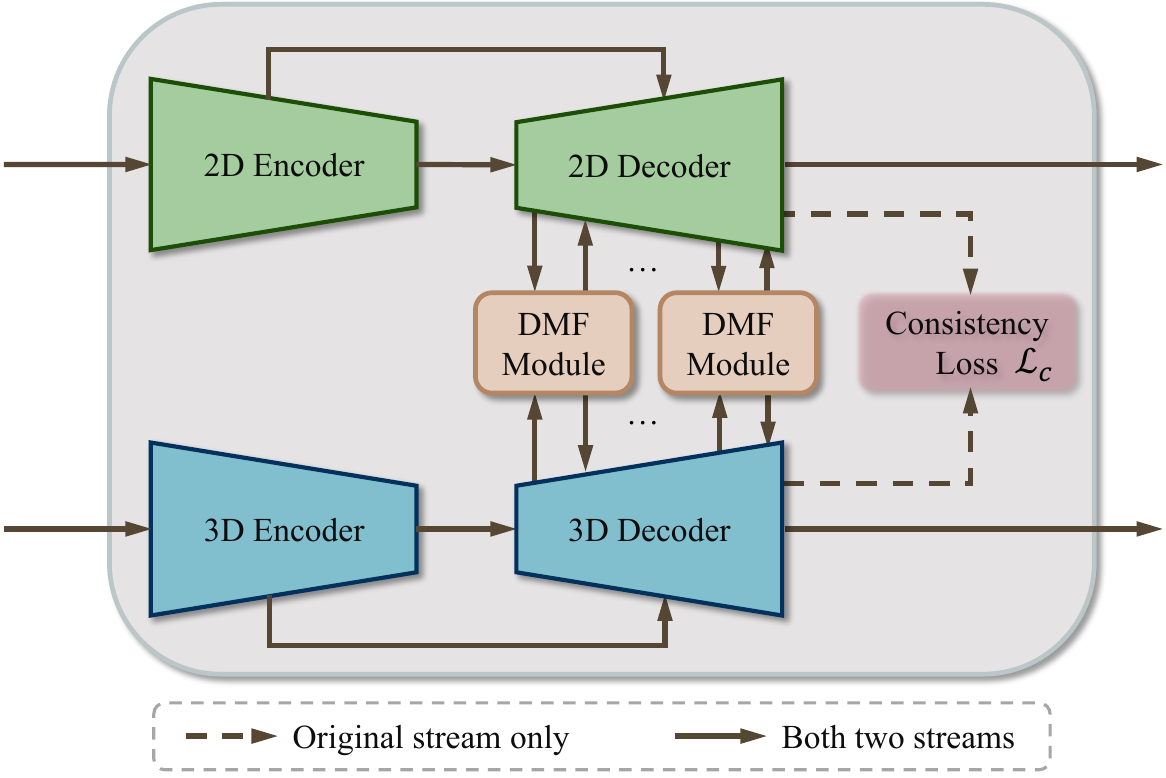}&\\
		\footnotesize (a) Architecture of the proposed PD-Net. & 
        \footnotesize (b) Architecture of the stream in PD-Net.\\
	\end{tabular}
	\caption{Architecture of the proposed PD-Net and the original stream / the pseudo-label prediction stream in PD-Net. The proposed PD-Net contains an original stream and a pseudo-label prediction stream. The Pseudo-label Optimization (PLO) module is utilized to optimize the pseudo labels output by the pseudo-label prediction stream. CE Loss represents the cross entropy loss. The labeled point clouds and their corresponding images are only trained in the original stream, while the unlabeled point clouds and their corresponding images are trained in both two streams. The stream in PD-Net contains 3D and 2D encoder-decoder branches for dual-modal data, and multiple Dual-modal Fusion (DMF) modules to fuse the dual-modal latent features. The consistency loss function is utilized to constrain the dual-modal output features in the original stream.}
	\label{fig: archi}
\end{figure*} 

In recent years, many methods \cite{2018splatnet, 20183dmv, 2019mvpnet, 2019upf, 2020xmuda, 2020fuseseg, 2021bpnet, Yan20222DPASS2P, Zhuang2021PerceptionAwareMF} have been proposed to jointly use the two modalities (i.e., 3D point clouds and images) to improve the semantic segmentation performances.
Dai \textit{et al.}  \cite{20183dmv} proposed to project the multi-view image features to the voxels and merge the multi-view features with the voxel features for better performance.
Considering the computational complexity of the voxel representation, Jaritz \textit{et al.}  \cite{2019mvpnet} designed a feature aggregation module to aggregate the 3D features projected from images to the original point clouds. 
Jaritz \textit{et al.}  \cite{2020xmuda} proposed to mutually project the sampled image and point cloud features, and minimize the distribution discrepancies between the dual-modal features.
Hu \textit{et al.}  \cite{2021bpnet} designed a bidirectional projection module where the point cloud and image features could interact with each other so that the advantages of these two modalities could be combined for better performance.
Based on \cite{2021bpnet}, Wang \textit{et al.} \cite{Wang2022SemAffiNetST} leveraged the semantic information to further enhance the mid-level features, which is proved to be helpful for improving both point cloud and image segmentation performances.
Zhuang \textit{et al.}  \cite{Zhuang2021PerceptionAwareMF} proposed a collaborative fusion scheme to exploit perceptual information from two modalities.
Yan \textit{et al.}  \cite{Yan20222DPASS2P} proposed a general training scheme to acquire semantic and structural information from the dual-modal data by distilling the information of 2D images to the 3D network.
Li \textit{et al.} \cite{Li_2023_CVPR_Mseg} proposed a method named MSeg3D. It utilizes joint intra-modal feature extraction and inter-modal feature fusion to mitigate the modality heterogeneity and explores the asymmetric multi-modal diversified augmentation transformations for effective training.

The above-mentioned fully-supervised methods require expensive cost for labeling, while the proposed PD-Net could simultaneously segment the point clouds and images with a small number of 3D labels and no 2D label needed.

\section{Methodology}
\label{sec: meth}
In this section, we introduce our proposed PD-Net in detail.
Firstly, we describe the overall architecture of the proposed network.
Then, we introduce the designed dual-modal fusion module, consistency loss function, and pseudo-label optimization module respectively.
Finally, we present the total loss function of the proposed network.
\subsection{Architecture}
The architecture of the proposed PD-Net is shown in Figure~\ref{fig: archi}a.
As seen from this figure, PD-Net employs a parallel two-stream structure: an original stream and a pseudo-label prediction stream.
The original stream is utilized to simultaneously segment the point clouds and images.
The pseudo-label prediction stream is utilized to predict pseudo labels for the unlabeled point clouds and their corresponding images for self-training in the original stream.
The parameters of the pseudo-label prediction stream $\mathbf{W}_{pl}$ are updated according to the original stream based on the EMA method \cite{2017meanteacher}.
The EMA method could retain the historical information via a progressive-update strategy, which could mitigate the negative influence brought by the false pseudo labels.
Specifically, the updated parameters of the original stream are denoted as $\mathbf{W}_{ori}'$ .
In the $s$-th training step, the updated parameters of the pseudo-label prediction stream $\mathbf{W}_{pl}'$ are formulated as:
\begin{equation}
		\mathbf{W}_{pl}' = \alpha \times \mathbf{W}_{pl} + (1 - \alpha) \times \mathbf{W}_{ori}',
	\label{equ:issnet_ema}
\end{equation}
where $\alpha = \min(1 - \frac{1}{s + 1}, t_{ema})$, and $t_{ema}$ is a predetermined threshold.
The labeled point clouds and their corresponding images are trained in the original stream, and the labels of images are projected from the labels of point clouds.
The unlabeled point clouds and their corresponding images are trained in both two streams, their corresponding outputs of the original stream are supervised by the pseudo labels output from the Pseudo-label Optimization (PLO) module.

In PD-Net, the original stream and the pseudo-label prediction stream have the same architecture, as shown in Figure~\ref{fig: archi}b. 
Both two streams contain a 3D encoder-decoder branch, a 2D encoder-decoder branch, and multiple Dual-modal Fusion (DMF) modules.
The 3D and 2D encoder-decoder branches are utilized to extract features from point clouds and images respectively.
\subsection{Dual-modal Fusion Module}
\begin{figure*}[ht]
	\centering
	\includegraphics[width=1\textwidth]{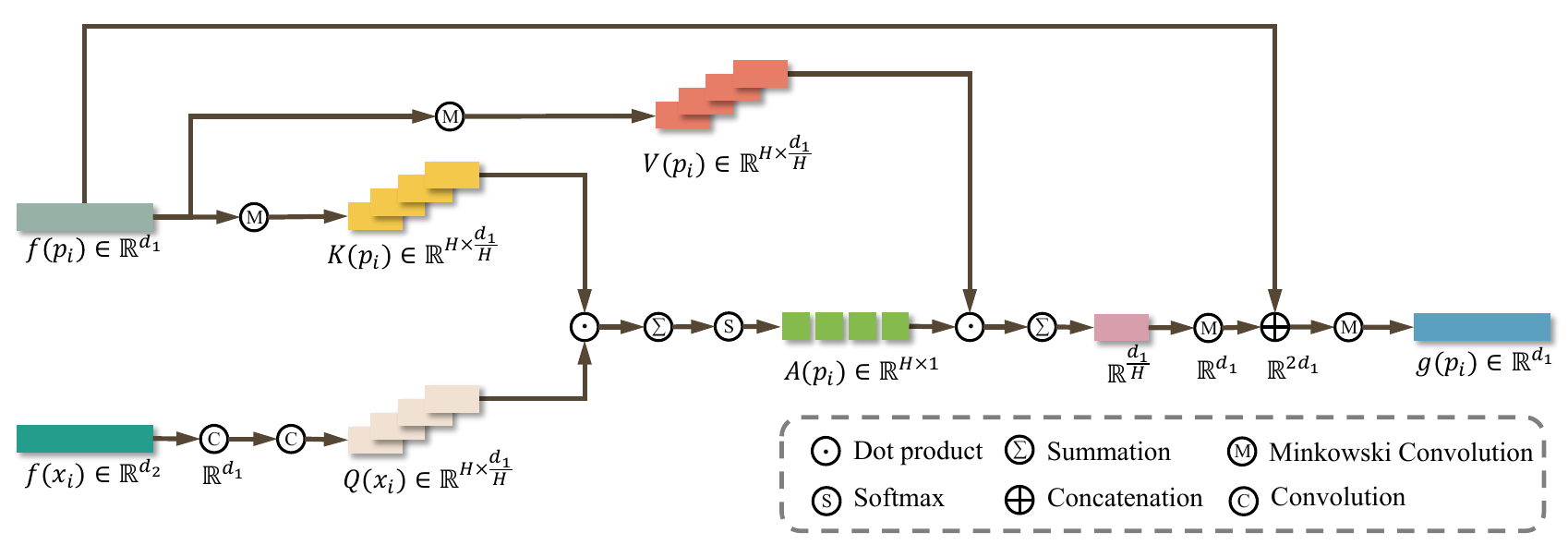}
	\caption{The calculation process of the 3D fused feature $\boldsymbol{g}(p_i)$ in the dual-modal fusion module. The dimensions of the key feature $K(\cdot)$, query feature $Q(\cdot)$, and value feature $V(\cdot)$ are the results of dividing the dimensions of their corresponding latent feature $\boldsymbol{f}(\cdot)$ by the head number $H$. $d_1$ and $d_2$ denote the dimensions of 3D features and 2D features respectively. The attention-based mechanism in the dual-modal fusion module facilitates adaptively learning complementary information from dual-modal data.}
	\label{fig:issnet_DMA}
\end{figure*}

The DMF module is used to fuse the latent features of point clouds and images at each layer of the 3D and 2D decoders.
The consistency loss term is explored to constrain the consistency of the 3D features and 2D features in the original stream.
The PLO module is utilized to optimize the coarse pseudo labels output by the pseudo-label prediction stream.

The Dual-modal Fusion (DMF) module is designed to fuse the 3D and 2D latent features. 
Considering the inherent domain gap between the two modalities, we only fuse the latent features of the paired points and pixels.
And the point-to-pixel correspondences could be easily calculated according to the pre-calibrated intrinsic and extrinsic parameters of the sensors.
The coordinates of the paired points and pixels are denoted as $\{p_i, x_i\}_{i=1}^{N}$, where $p_i \in \mathbb{R}^3$ is the coordinate of the point, $x_i \in \mathbb{Z}^ 2$ is the coordinate of the pixel, and $N$ is the number of matching pairs. The DMF module takes the paired 3D feature $\boldsymbol{f}(p_i)$ and 2D feature $\boldsymbol{f}(x_i)$ from the current 3D and 2D decoder layers as input, and outputs the fused 3D feature $\boldsymbol{g}(p_i)$ and fused 2D feature $\boldsymbol{g}(x_i)$, which are further fed into the next 3D and 2D decoder layers, respectively.

\textbf{Learning 3D fused features:}
Multi-head attention-based fusion mechanism is employed to fuse the paired latent features in the DMF module. Figure~\ref{fig:issnet_DMA} illustrates the calculation process of the 3D fused feature $\boldsymbol{g}(p_i)$. 
Specifically, Minkowski convolution \cite{2019minkowskinet} operation and convolution operation are performed on $\boldsymbol{f}(p_i)$ and $\boldsymbol{f}(x_i)$ respectively to extract their corresponding key feature $K(\cdot)$, query feature $Q(\cdot)$, and value feature $V(\cdot)$.
Then, the dot product, summation, and Softmax operations are performed on the 3D key feature $K(p_i)$ and 2D query feature $Q(x_i)$ to obtain the 3D attention map $A(p_i)$. Weighted summation is performed on $A(p_i)$ and 3D value feature $V(p_i)$ to obtain the multi-head attention feature, which is extended to the same dimension with $\boldsymbol{f}(p_i)$ by a Minkowski convolution layer and concatenated with $\boldsymbol{f}(p_i)$. Finally, the concatenated feature passes through a Minkowski convolution layer to output the 3D fused feature.

The above-mentioned calculation process of the 3D fused feature $\boldsymbol{g}(p_i)$ could be formulated as:
\begin{equation}
		\boldsymbol{g}(p_i) =\textup{M}\bigg(\textup{M}\Big(\sum\big(A(p_i) \odot V(p_i)\big)\Big) 
		\oplus \boldsymbol{f}(p_i)\bigg),
\label{equ:issnet_DMA_1}
\end{equation}
where $A(p_i)=\textup{Softmax}\Big(\sum \big(K(p_i ) \odot 
 Q(x_i)\big)\Big)$, $\textbf{M}$ denotes the Minkowski convolution, $\odot$ denotes the dot product, and $\oplus$ denotes the concatenation.

\textbf{Learning 2D fused features:}
Similarly, the calculation process of the 2D fused feature $\boldsymbol{g}(x_i)$ is formulated as:
\begin{equation}
		\boldsymbol{g}(x_i) =\textup{C}\bigg(\textup{C}\Big(\sum\big(A(x_i) \odot V(x_i)\big)\Big) 
		\oplus \boldsymbol{f}(x_i)\bigg),
\label{equ:issnet_DMA_2}
\end{equation}
where $A(x_i)=\textup{Softmax}\Big(\sum\big(K(x_i ) \odot 
 Q(p_i)\big)\Big)$, and $\textbf{C}$ denotes the commonly-used convolution.

We utilize multiple DMF modules to fuse the 3D and 2D latent features in multiple scales.
Compared with direct feature concatenation, the proposed multi-head attention-based fusion mechanism could extract more discriminative and informative features from the two modalities,  which would be demonstrated in Subsection~\ref{ablation}.
\subsection{Consistency Loss}\label{subsection:clf}

In order to constrain the consistency between the learned dual-modal features in the output feature spaces of the original stream, we propose the consistency loss term.

The paired output features of the 3D and 2D branches in the original stream are denoted as $\{\boldsymbol{y}(p_i), \boldsymbol{y}(x_i)\}_{i=1}^N$.
Then, the proposed consistency loss term $Loss_{c}$ is formulated as:
\begin{equation}
	\begin{split}
		&\mathcal{L}_{c} = \frac{1}{N}\sum_{i=1}^N||\boldsymbol{y}(p_i) - \boldsymbol{y}(x_i)||_2^2,
	\end{split}
	\label{equ:issnet_mse}
\end{equation}
where $||\cdot||$ denotes the L2-norm.

Overall, the dual-modal fusion module and the consistency loss term fuse the dual-modal features in different levels of feature space.
\subsection{Pseudo-label Optimization Module}
\begin{figure}[b]
	\centering
	\includegraphics[width=1\linewidth]{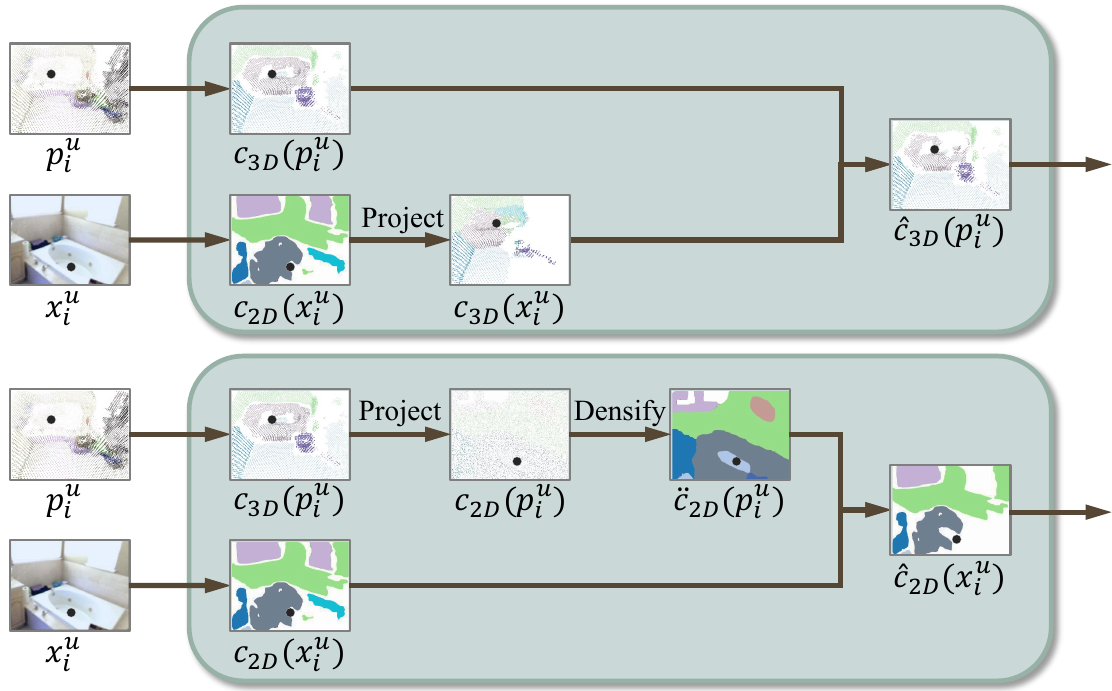}
	\caption{The optimization process of 3D (top) and 2D (bottom) pseudo-labels. The coarse 2D pseudo labels are projected to point clouds to obtain the projected 3D pseudo labels. The coarse 3D pseudo labels are densified after being projected to the image plane to obtain the projected 2D pseudo labels. The black point denotes the pseudo label that is deleted by the pseudo-label optimization module.}
	\label{fig:issnet_plf}
\end{figure}
Due to the inherent limitations of the two modalities (i.e., the lack of texture information in point clouds and the lack of depth information in images), the 3D and 2D encoder-decoder branches tend to predict pseudo labels for objects according to their geometric structures and textures respectively.

In order to guarantee the effectiveness of the self-training of unlabeled data in the original stream, we propose the Pseudo-label Optimization (PLO) module based on a voting mechanism to improve the reliability of the pseudo labels.
The PLO module is only utilized for the paired unlabeled points and pixels, whose coordinates are denoted as $\{p_i^u, x_i^u\}_{i=1}^{N_u}$, where $N_u$ is the number of the unlabeled matching pairs.
It takes the coarse pseudo labels of the unlabeled points and pixels as input and outputs their corresponding optimized pseudo labels.

Specifically, the coarse 3D and 2D pseudo labels of the paired unlabeled point $p_i^u$ and pixel $x_i^u$ , which are output by the pseudo-label prediction stream, are denoted as  $c_{3D}(p_i^u)$ and $c_{2D}(x_i^u)$ respectively.
And the process of optimizing 3D (top) and 2D (bottom) pseudo labels by the PLO module is shown in Figure~\ref{fig:issnet_plf}.

\textbf{Optimization of 3D pseudo labels: }Firstly, the coarse 2D pseudo label $c_{2D}(x_i^u)$ is projected to its paired point to obtain the projected 3D pseudo label $c_{3D}(x_i^u)$. 
The coarse 3D pseudo label $c_{3D}(p_i^u)$ is retained if it is consistent with $c_{3D}(x_i^u)$. 
Otherwise,  a confidence-based filtering mechanism is utilized.
The confidence of the coarse 3D pseudo label is simply the value of the $c_{3D}(p_i^u)$-th dimension of the 3D output feature in the pseudo-label prediction stream, which is denoted as $\gamma_{c}(p_i^u, c_{3D}(p_i^u))$.
The coarse 3D pseudo label  $c_{3D}(p_i^u)$ is retained if its confidence is larger than the predetermined confidence threshold $t_{conf}$.
Otherwise, the coarse 3D pseudo label is deleted.
The above process could be formulated as:
\begin{equation}
	\begin{aligned}
		\hat{c}_{3D}(p^u_i) =
		\begin{cases}
			c_{3D}(p^u_i)
			&\text{, }c_{3D}(x^u_i) = c_{3D}(p^u_i)\text{ or }\\
   &\gamma_{c}(p^u_i, c_{3D}(p^u_i)) > t_{conf},\\
			\text{deleted }
			& \text{, other, }
		\end{cases}
	\end{aligned}
	\label{equ:issnet_plf_1}
\end{equation}
where $\hat{c}_{3D}(p^u_i)$ is the optimized 3D pseudo label.

\textbf{Optimization of 2D pseudo labels:} 
As seen in the bottom part of Figure~\ref{fig:issnet_plf}, the projected 2D pseudo label $c_{2D}(p_i^u)$ is sparse, which leaves the majority of the pixels unprojected.
To address this issue, we project each 3D output feature in the pseudo-label stream into the image plane and perform average pooling in the local areas for the unprojected pixels.
The dimension with the largest value in the pooled output feature is selected as the dense 2D pseudo label $\ddot{c}_{2D}(p_i^u)$.

Similarly, the optimization process of the 2D pseudo label is formulated as:
\begin{equation}
	\begin{aligned}
		\hat{c}_{2D}(x^u_i) =
		\begin{cases}
			c_{2D}(x^u_i)
			&\text{, }\ddot{c}_{2D}(p^u_i)=c_{2D}(x^u_i)\text{ or}\\
   &\gamma_{c}(x^u_i, c_{2D}(x^u_i))>t_{conf},\\
			\text{deleted }
			& \text{, other,}
		\end{cases}
	\end{aligned}
	\label{equ:issnet_plf_2}
\end{equation}
where $\hat{c}_{xD}(x^u_i)$ is the optimized 2D pseudo label.

\subsection{Total Loss Function}

As depicted in Figure~\ref{fig: archi}a, four cross-entropy loss terms are employed for the labeled point clouds and their corresponding images, and the unlabeled point clouds and their corresponding images, which are denoted as $\mathcal{L}_{3D}^l$, $\mathcal{L}_{2D}^l$, $\mathcal{L}_{3D}^u$, and $\mathcal{L}_{2D}^u$. 
And their targets are the ground truth 3D labels, projected 2D labels, optimized 3D pseudo labels, and optimized 2D pseudo labels, respectively.
Combined with the consistency loss term $\mathcal{L}_{c}$ in Subsection~\ref{subsection:clf}, the total loss function $\mathcal{L}_{total}$ is formulated as:
\begin{equation}
		\mathcal{L}_{total} = \mathcal{L}_{3D}^l+\mathcal{L}_{2D}^l+\mathcal{L}_{3D}^u+\mathcal{L}_{2D}^u+\lambda_{c}\mathcal{L}_{c},
	\label{equ:issnet_loss_all}
\end{equation}
where $\lambda_{c}$ is the weight of the consistency loss term.

\section{Experiments}
\label{sec: exp}

\subsection{Experimental Setup}
\begin{table*}[t]
	\begin{center}
		\caption{Evaluation results on the validation set of the ScanNet dataset \cite{2017scannet}. The best results are in bold in each metric.}
		\label{tab:issnet_scannet_1}
			\begin{tabular}{c|lcccccc}
				\hline
				&&\multicolumn{3}{c}{Point Cloud}	&\multicolumn{3}{c}{Image}\\
				\rule{0pt}{8pt}&Method	&mIoU	&mAcc	&OA	&mIoU	&mAcc	&OA\\
				\hline
				\multirow{9}{*}{20\%}
				\rule{0pt}{8pt}&MinkowskiNet18A \cite{2019minkowskinet} &59.31	&67.92	&84.13	&-&-&-	\\
				\rule{0pt}{8pt}&ResNet34 \cite{2016resnet} &-&-&-&45.04	&59.20	&74.96	\\	
				\cline{2-8}
				\rule{0pt}{8pt}&Deng \textit{et al.}  \cite{2022sssnet}&55.12	&63.61	&82.43	&-&-&-\\
				\rule{0pt}{8pt}&TCSM-V2 \cite{2020tcsmv2}&-&-&-	&52.65	&61.08	&80.17\\
				\rule{0pt}{8pt}&CPS \cite{2021cps}&-&-&- &55.23	&64.97	&81.86\\
				\rule{0pt}{8pt}&$\pi$-Model \cite{2017pimodel} &60.41	&69.08	&84.34	&51.09	&60.08	&78.81	\\
				\rule{0pt}{8pt}&Mean Teacher \cite{2017meanteacher} &61.12	&69.47	&84.72	&51.82	&60.56	&79.68	\\
				\rule{0pt}{8pt}&Pseudo-Labels \cite{2013pseudolabels} &60.64	&69.27	&84.51	&53.01	&62.17	&80.68	\\
				\cline{2-8}
				\rule{0pt}{8pt}&PD-Net &\textbf{63.38}	&\textbf{71.61}	&\textbf{86.28}	&\textbf{60.17}	&\textbf{70.78}	&\textbf{83.29}	\\
				\hline
				\multirow{9}{*}{10\%}
				\rule{0pt}{8pt}&MinkowskiNet18A \cite{2019minkowskinet} &52.27	&61.19	&80.81	&-&-&-	\\
				\rule{0pt}{8pt}&ResNet34 \cite{2016resnet} &-&-&-&43.00	&55.09	&72.75	\\	
				\cline{2-8}
				\rule{0pt}{8pt}&Deng \textit{et al.}  \cite{2022sssnet}&52.38	&60.76	&81.18	&-&-&-\\
				\rule{0pt}{8pt}&TCSM-V2 \cite{2020tcsmv2}&-&-&-	&46.98	&57.61	&74.25\\
				\rule{0pt}{8pt}&CPS \cite{2021cps}&-&-&- &48.01&58.76&76.02\\
				\rule{0pt}{8pt}&$\pi$-Model \cite{2017pimodel} &54.84	&63.45	&81.54	&47.72	&57.56	&75.57	\\
				\rule{0pt}{8pt}&Mean Teacher \cite{2017meanteacher} &55.24	&63.70	&81.79	&46.63	&57.39	&74.10	\\
				\rule{0pt}{8pt}&Pseudo-Labels \cite{2013pseudolabels} &54.47	&63.38	&81.40	&46.83	&57.51	&74.12	\\
				\cline{2-8}
				\rule{0pt}{8pt}&PD-Net &\textbf{58.38}	&\textbf{67.23}	&\textbf{83.68}	&\textbf{50.80}	&\textbf{60.77}	&\textbf{79.38}	\\
				\hline
			\end{tabular}
	\end{center}
\end{table*}
\textbf{Dataset:}
We evaluate the proposed PD-Net on the ScanNet dataset  \cite{2017scannet}, which contains 1613 indoor point clouds reconstructed from depth images. 
In addition, the ScanNet dataset contains more than $2.5 \times 10^6$ RGB images, and each point cloud corresponds to more than 5000 images.
The intrinsic and extrinsic parameters of the sensors are also provided, which enables the calculation of the point-to-pixel correspondences.
Both the 3D point clouds and images in the ScanNet dataset are annotated with 20 semantic categories.

\textbf{Implementation details and metrics:}
In this work, the 3D and 2D encoder-decoder branches,  which are based on MinkowskiNet18A \cite{2019minkowskinet} and ResNet34 \cite{2016resnet} respectively, both use the U-Net \cite{2015unet} architectures.  
For each 3D point cloud, we randomly sample 3 images from its corresponding image set for dual-modal training.
The weight threshold $t_{ema}$ in the EMA method is set to 0.999, 
the head number in the dual-modal fusion module is set to 4,
the confidence threshold $t_{conf}$ in the pseudo-label optimization module is set to 0.9 for deleting the false labels with low confidences,
and the consistency loss weight $\lambda_{c}$ is set to 5.
The voxel size is set to 5cm for efficient training.
We apply the Stochastic Gradient Descent (SGD) optimizer with a base learning rate of 0.01. The batch size and epoch number is set as 16 and 150 respectively.

For evaluating the performance of semi-supervised segmentation, we split the training point clouds into a labeled set and an unlabeled set.
Specifically, we randomly sample the labeled point clouds from the training point clouds with two different ratios (i.e., 20\% and 10\%).
Only the labeled point clouds and their corresponding images are trained in the first 100 epochs for a more stable semi-supervised training.
The unlabeled point clouds and their corresponding images are incorporated into training in the last 50 epochs.

We use mean Intersection over Union (mIoU), mean Accuracy (mAcc), and Overall Accuracy (OA) as the evaluation metrics for both 3D and 2D semantic segmentation.

\subsection{Comparative Evaluation}
Considering the 3D and 2D encoder-decoder branches are based on MinkowskiNet18A \cite{2019minkowskinet} and ResNet34 \cite{2016resnet} respectively, we evaluate the performances of the baseline models (i.e., MinkowskiNet18A and ResNet34) by training on the labeled set. 
Then, we compare the proposed PD-Net with several semi-supervised uni-modal semantic segmentation methods for point clouds \cite{2022sssnet} and images \cite{2020tcsmv2,2021cps}.
\begin{table}[!b]
	\begin{center}
		\caption{Evaluation results on the validation set of the ScanNet. * denotes that point clouds and images are trained jointly. \dag$\;$denotes that only depth images are used for training.  \S$\;$ denotes that only RGB images are used for training. \# denotes that RGB-D images are used for training.}
		\label{tab:issnet_scannet_test}
			\begin{tabular}{lc|lc}
				\hline
				Method	&3D mIoU &Method	&2D mIoU\\
				\hline 
				Pointnet++ \cite{2017pointnet++} &53.5 &ERFNetEnc \S \cite{Romera2018ERFNetER} &51.7\\
                PoinConv \cite{2019pointconv}&61.0 &AdaptNet++ \S \cite{2018adapt} &52.9\\
                PointASNL \cite{2020pointasnl}&63.5 &AdaptNet++ \dag \cite{2018adapt}&53.8\\
                MVPNet \cite{2019mvpnet}&65.0  &Deeplabv3 \S \cite{deeplabv3} &56.1\\
				Minkowski42 \cite{2019minkowskinet}&68.0&ERFNetEnc \dag \cite{Romera2018ERFNetER}&56.7 \\
				KPConv \cite{2019kpconv}&69.2 &SSMA \# \cite{Valada2018SelfSupervisedMA} &61.1\\
    			JointPointBased \cite{2019upf}&69.2 &RFBNet \# \cite{Deng2019RFBNetDM}&62.6\\
                PointTransformer \cite{2021pointtransformer}&70.6&GRBNet \# \cite{Qian2022GatedResidualBF}&62.6\\
				BPNet * \cite{2021bpnet}&73.9& MCA-Net \# \cite{Shi2020MultilevelCR}&64.3\\
                StratifiedPT \cite{Lai2022StratifiedTF}&74.3&BPNet * \cite{2021bpnet}&71.9\\
				\hline
				PD-Net * (20\%) &63.4 &PD-Net * (20\%) &60.2\\
				PD-Net * (10\%) &58.4 &PD-Net * (10\%) &50.8\\
				\hline
			\end{tabular}
	\end{center}
\end{table}
In addition, several typical semi-supervised learning methods are extended to tackle the semi-supervised dual-modal semantic segmentation task, including $\pi$-Model \cite{2017pimodel}, Mean Teacher \cite{ 2017meanteacher}, and Pseudo-Labels \cite{2013pseudolabels}. 
We evaluate these semi-supervised learning methods based on MinkowskiNet18A and ResNet34 while retaining their other experimental settings for a fair comparison. 
All these comparative methods utilize the same labeled set and unlabeled set. 
\begin{figure*}[!t]
	\centering
	\includegraphics[width=1.00\textwidth]{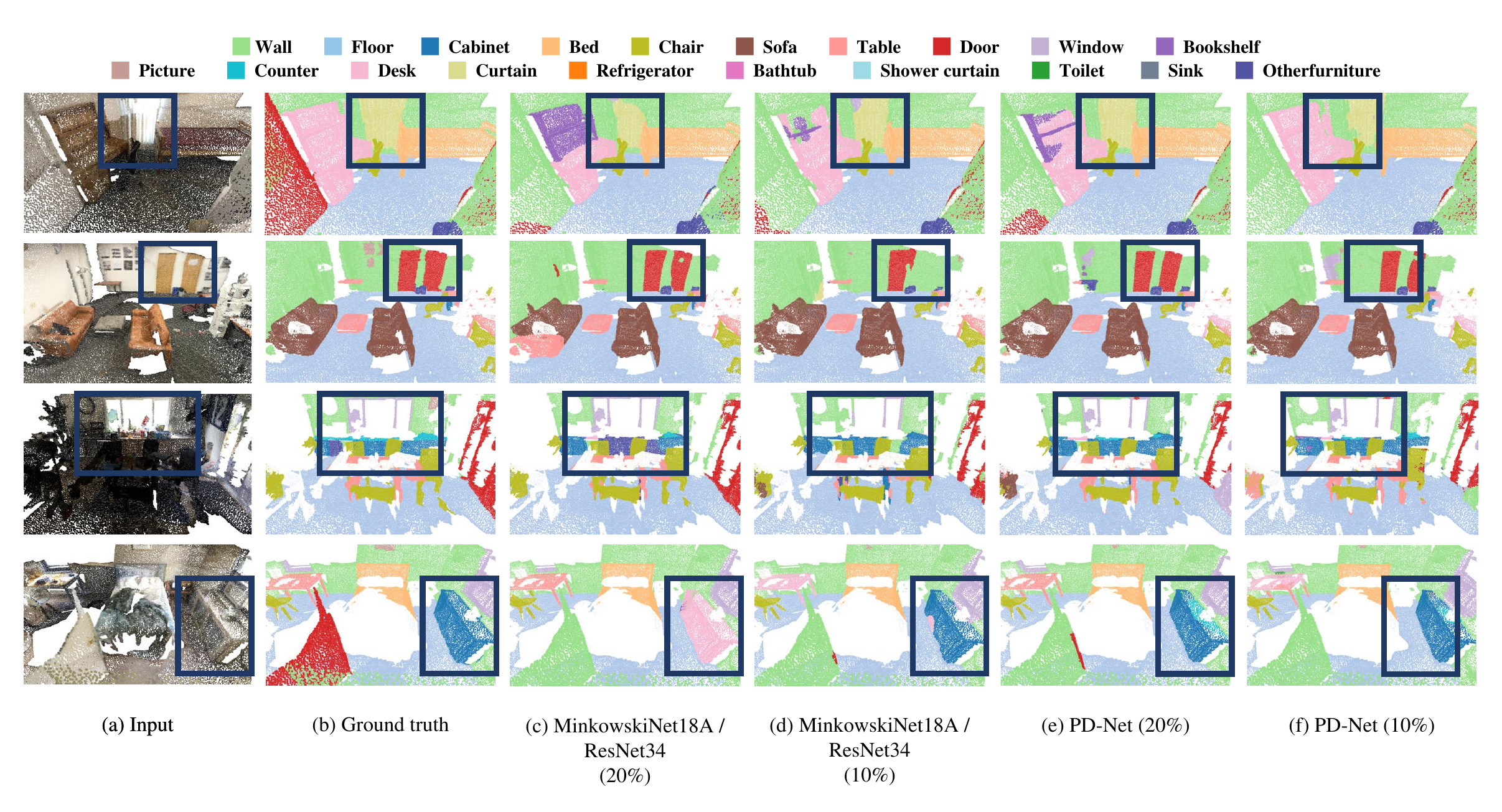}
	\caption{Qualitative results of point cloud segmentation on the validation set of the ScanNet \cite{2017scannet}. The segmentation results of the baseline model (MinkowskiNet18A \cite{2019minkowskinet} and ResNet34 \cite{2016resnet}) and our proposed PD-Net in two different labeled-ratio settings (20\% and 10\%) are visualized.}
	\label{pointres}
\end{figure*}
\begin{figure*}[!t]
	\centering
	\includegraphics[width=1.00\textwidth]{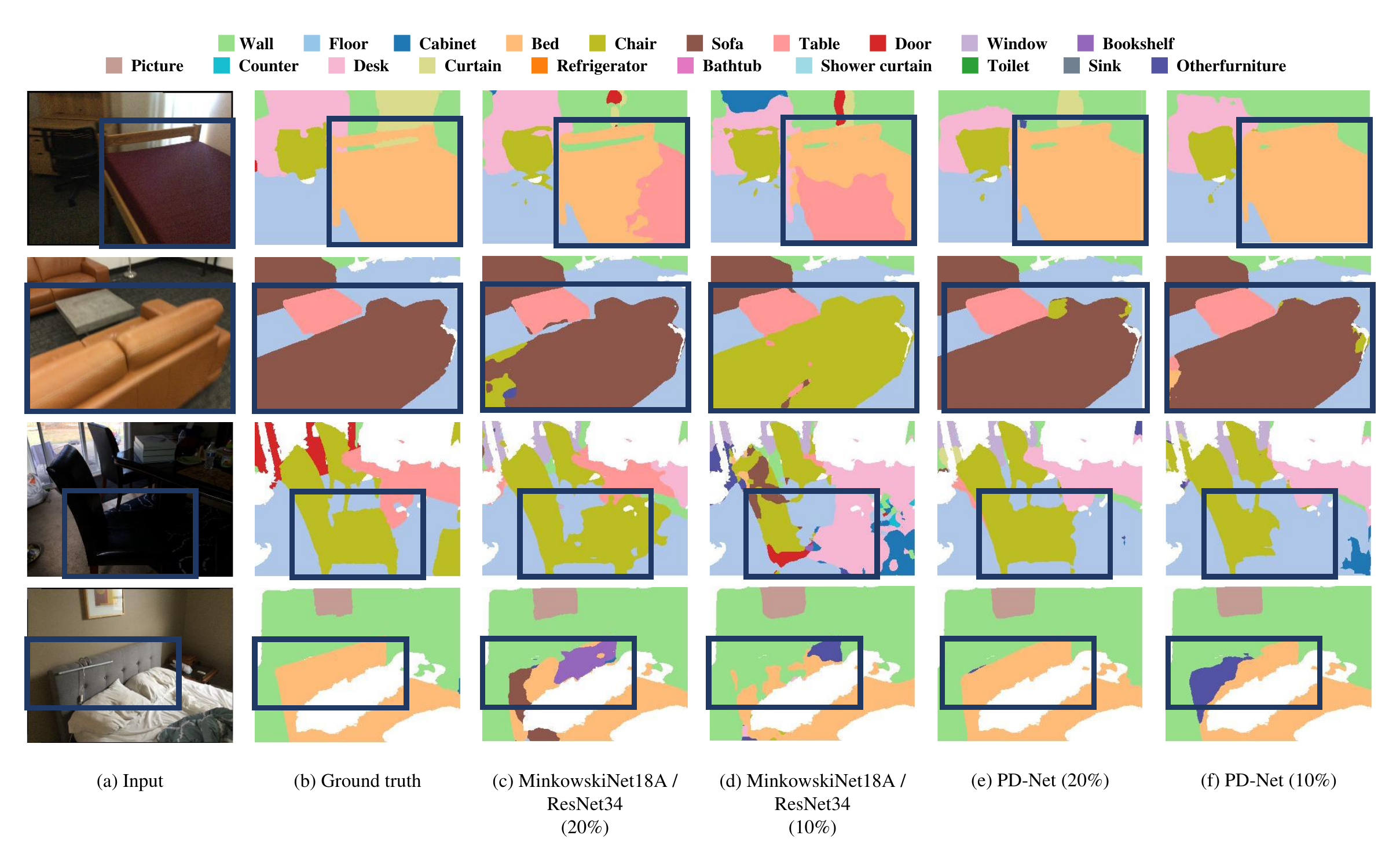}
	\caption{Qualitative results of image segmentation on the validation set of the ScanNet \cite{2017scannet}. The segmentation results of the baseline model (MinkowskiNet18A \cite{2019minkowskinet} and ResNet34 \cite{2016resnet}) and our proposed PD-Net in two different labeled-ratio settings (20\% and 10\%) are visualized.}
	\label{imageres}
\end{figure*}
Table~\ref{tab:issnet_scannet_1} reports the quantitative results of the proposed PD-Net and comparative methods on the validation set of the ScanNet \cite{2017scannet}.
As seen from this table, in two different labeled-ratio settings, the proposed PD-Net outperforms all the comparative methods in point cloud segmentation and image segmentation tasks.
The proposed PD-Net outperforms Pseudo-Labels, Deng \textit{et al.}  \cite{2022sssnet}, TCSM-V2 \cite{2020tcsmv2}, and CPS \cite{2021cps}, 
because it could effectively utilize the complementary information from the point clouds and images, and mitigate the negative impact brought by the falsely predicted pseudo labels to some extent.
And the proposed PD-Net outperforms the $\pi$-Model and Mean Teacher, 
probably because the consistency constraints between the latent features from different scales are more powerful than the constraints between the output features from different transformations.

Figure~\ref{pointres} and Figure~\ref{imageres} visualize the qualitative results of point cloud segmentation and image segmentation respectively. As seen from these figures, the proposed PD-Net predicts more accurately than MinkowskiNet18A and ResNet34 in both two labeled-ratio settings. 
We highlight the key regions in the dark blue boxes.
The visualization results demonstrate that the proposed PD-Net yields promising performances in 3D and 2D segmentation with only a small number of labeled point clouds required.

In addition, we compare the proposed PD-Net, which is trained in a semi-supervised manner, with several typical fully-supervised semantic segmentation methods for point clouds \cite{2017pointnet++, 2019pointconv,2020pointasnl,2019mvpnet,2019minkowskinet,2019kpconv,2019upf,2021pointtransformer,2021bpnet,Lai2022StratifiedTF} and for images \cite{Romera2018ERFNetER,2018adapt,deeplabv3,Valada2018SelfSupervisedMA,Deng2019RFBNetDM,Qian2022GatedResidualBF,Shi2020MultilevelCR,2021bpnet} on the validation set of the ScanNet. 
The corresponding results are reported in Table~\ref{tab:issnet_scannet_test}.
As seen from this table, the PD-Net trained under the 20\%-labeled setting achieves comparable results with the comparative fully-supervised methods, which further demonstrates the effectiveness of the proposed PD-Net.

\subsection{Ablation Study}
\label{ablation}
\begin{table}[t]
	\begin{center}
		\caption{Ablation studies of the involved components.}
		\label{tab:issnet_module}
			\begin{tabular}{c|lcccccc}
				\hline
				&&\multicolumn{3}{c}{Point Cloud}	&\multicolumn{3}{c}{Image}\\
				\rule{0pt}{8pt}&Model	&mIoU	&mAcc	&OA	&mIoU	&mAcc	&OA\\
				\hline
				\multirow{5}{*}{20\%}
				\rule{0pt}{8pt}&Baseline &59.31	&67.92	&84.13	&45.04	&59.20	&74.96	\\	
				\rule{0pt}{8pt}&Model A &60.64	&69.27	&84.51	&53.01	&62.17	&80.68	\\
				\rule{0pt}{8pt}&Model B &61.46	&70.20	&85.03	&55.02	&64.59	&80.86	\\
				\rule{0pt}{8pt}&Model C &62.26  &\textbf{72.11}  &85.56  &58.30  &69.83  &82.56  \\
				\rule{0pt}{8pt}&PD-Net &\textbf{63.38}	&71.61	&\textbf{86.28}	&\textbf{60.17}	&\textbf{70.78}	&\textbf{83.29}	\\	
				\hline
				\multirow{5}{*}{10\%}
				\rule{0pt}{8pt}&Baseline &52.27	&61.19	&80.81	&43.00	&55.09	&72.75	\\	
				\rule{0pt}{8pt}&Model A &54.47	&63.38	&81.40	&46.83	&57.51	&74.12	\\
				\rule{0pt}{8pt}&Model B &54.72	&67.66	&81.52	&47.07	&57.94	&75.34	\\
				\rule{0pt}{8pt}&Model C &56.87	&66.49	&82.90	&50.14	&59.61	&79.18	\\
				\rule{0pt}{8pt}&PD-Net &\textbf{58.38}	&\textbf{67.23}	&\textbf{83.68}	&\textbf{50.80}	&\textbf{60.77}	&\textbf{79.38}	\\		
				\hline
			\end{tabular}
	\end{center}
\end{table}
\begin{table}[t]
	\begin{center}
		\caption{Results of PD-Net with different fusion modules.}
		\label{tab:issnet_DMA}
			\begin{tabular}{c|lcccccc}
				\hline
				&&\multicolumn{3}{c}{Point Cloud}	&\multicolumn{3}{c}{Image}\\
				\rule{0pt}{8pt}&Module	&mIoU	&mAcc	&OA	&mIoU	&mAcc	&OA\\
				\hline
				\multirow{2}{*}{20\%}
				\rule{0pt}{8pt}&BP \cite{2021bpnet}&62.63	&71.04	&85.80	&58.40	&69.81	&82.10	\\
				\rule{0pt}{8pt}&DMF &\textbf{63.38}	&\textbf{71.61}	&\textbf{86.28}	&\textbf{60.17}	&\textbf{70.78}	&\textbf{83.29}	\\	
				\hline
				\multirow{2}{*}{10\%}
				\rule{0pt}{8pt}&BP \cite{2021bpnet} &57.19	&66.61	&83.01	&50.20	&59.74	&79.02	\\
				\rule{0pt}{8pt}&DMF &\textbf{58.38}	&\textbf{67.23}	&\textbf{83.68}	&\textbf{50.80}	&\textbf{60.77}	&\textbf{79.38}	\\	
				\hline
			\end{tabular}
	\end{center}
\end{table}
The effectiveness of each key element in the proposed PD-Net is verified by conducting ablation studies on the validation set of ScanNet dataset \cite{2017scannet}. 
The following models under two labeled-ratio settings are compared:
\begin{itemize}
\item Baseline: The 3D and 2D encoder-decoder branches (based on MinkowskiNet18A \cite{2019minkowskinet} and ResNet34 \cite{2016resnet}) trained on the labeled point clouds and their corresponding images.
\item Model A: Based on Baseline, the pseudo-label supervision for unlabeled data is added.
\item Model B: Based on Model A, the EMA method \cite{2017meanteacher} is utilized to update the parameters of the pseudo-label prediction stream.
\item Model C: Based on Model B, the Pseudo-label Optimization (PLO) module and the consistency loss term are added.
\item PD-Net (the whole model): Based on Model C, the Dual-modal Fusion (DMF) module is added.
\end{itemize}

The corresponding results are reported in Table~\ref{tab:issnet_module}. 
As seen from this table, Model A performs better than the Baseline, indicating that the coarse pseudo labels generated by the pseudo-label prediction stream could supervise the unlabeled data to some extent. 
Model B makes further progress based on Model A, demonstrating that using the historical information to update the parameters of the pseudo-label prediction stream is superior to the common updating strategy.
The performances of Model C are promoted based on Model B, which is attributed to the consistency constraints between the 3D and 2D output features and the optimization of the pseudo labels.
The whole PD-Net achieves the best results in most cases, probably because the DMF module could effectively fuse the dual-modal latent features.

To further verify the superiority of the DMF module, we replace the DMF module with a similar module for dual-modal feature fusion, while keeping the experimental settings and other modules unchanged. Specifically, we choose the Bidirectional Projection (BP) module in BPNet \cite{2021bpnet}.

The comparison results are reported in Table~\ref{tab:issnet_DMA}.
As seen from this table, the model with DMF module achieves better segmentation performances, indicating that the multi-head attention-based mechanism has stronger fusion ability than the view fusion strategy in the BP module which simply learns the impact factors for each view at every point.


\begin{table}[b]
	\begin{center}
		\caption{Results of PD-Net with different $t_{conf}$.}
		\label{tab:issnet_threshold}
			\begin{tabular}{c|ccccccc}
				\hline
				&&\multicolumn{3}{c}{Point Cloud}	&\multicolumn{3}{c}{Image}\\
				\rule{0pt}{8pt}&$t_{conf}$	&mIoU	&mAcc	&OA	&mIoU	&mAcc	&OA\\
				\hline
				\multirow{4}{*}{20\%}
                \rule{0pt}{8pt}&0.60&62.31	&70.35	&85.42	&58.37	&68.66	&81.47	\\	
				\rule{0pt}{8pt}&0.85&62.83	&71.04	&85.72	&58.89	&69.12	&82.45	\\	
				\rule{0pt}{8pt}&0.90&\textbf{63.38}	&\textbf{71.61}	&\textbf{86.28}	&\textbf{60.17}	&\textbf{70.78}	&\textbf{83.29}	\\	
				\rule{0pt}{8pt}&0.95&62.95	&71.23	&85.95	&59.10	&69.30	&82.78	\\	
				\hline
				\multirow{4}{*}{10\%}
                \rule{0pt}{8pt}&0.60&56.14	&65.87	&82.36	&48.79	&58.61	&78.05	\\	
				\rule{0pt}{8pt}&0.85&57.47	&66.62	&83.14	&49.30	&58.92	&78.72	\\
				\rule{0pt}{8pt}&0.90&\textbf{58.38}	&\textbf{67.23}	&\textbf{83.68}	&\textbf{50.80}	&\textbf{60.77}	&\textbf{79.38}	\\
				\rule{0pt}{8pt}&0.95&57.55	&66.82	&83.18	&50.24	&60.22	&78.98	\\	
				\hline
			\end{tabular}
	\end{center}
\end{table}

\subsection{Analysis on Hyper-parameters}

In this section, we provide more analysis on some hyper-parameters, including the confidence threshold $t_{conf}$ in the pseudo-label optimization module, the weight of consistency loss $\lambda_{c}$, and the voxel size. The experiments are conducted on the validation set of ScanNet \cite{2017scannet}.

\textbf{Effect of confidence threshold.} As seen in \eqref{equ:issnet_plf_1} and \eqref{equ:issnet_plf_2}, the confidence threshold $t_{conf}$ affects the quality of the optimized pseudo labels. 
We evaluate the proposed PD-Net with $t_{conf}=\{0.6, 0.85, 0.9,0.95\}$ to estimate the insensitive range of $t_{conf}$.
The corresponding results are reported in Table \ref{tab:issnet_threshold}, which indicate that our model achieves relatively stable performances when $t_{conf}$ ranges in [0.85, 0.95] and $t_{conf}$ with a lower value (i.e., $t_{conf}=\{0.6\}$) may impair the quality of pseudo labels, and thus deteriorate the performances.
The model achieves the best performances when $t_{conf}=0.9$.

\begin{table}[t]
	\begin{center}
		\caption{Results of PD-Net with different $\lambda_{c}$.}
		\label{tab:issnet_loss_weight}
			\begin{tabular}{c|ccccccc}
				\hline
				&&\multicolumn{3}{c}{Point Cloud}	&\multicolumn{3}{c}{Image}\\
				\rule{0pt}{8pt}&$\lambda_{c}$	&mIoU	&mAcc	&OA	&mIoU	&mAcc	&OA\\
				\hline
				\multirow{6}{*}{20\%}
                \rule{0pt}{8pt}&0&61.72	&70.65	&84.87	&55.28	&64.63	&80.52	\\
				\rule{0pt}{8pt}&0.2&62.91	&71.14	&85.96	&56.86	&67.82	&81.08	\\
				\rule{0pt}{8pt}&1&63.21	&71.42	&86.08	&59.31	&69.53	&83.14	\\
				\rule{0pt}{8pt}&5&\textbf{63.38}	&\textbf{71.61}	&\textbf{86.28}	&\textbf{60.17}	&\textbf{70.78}	&83.29	\\	
				\rule{0pt}{8pt}&10&62.82  &70.92  &85.99  &59.17  &68.82  &\textbf{83.66}  \\
                \rule{0pt}{8pt}&50&61.98&69.84	&84.31	&56.50	&64.92	&80.69\\
				\hline
				\multirow{6}{*}{10\%}
                \rule{0pt}{8pt}&0&55.11	&66.07	&81.96	&47.56	&57.08	&75.69	\\
				\rule{0pt}{8pt}&0.2&57.90	&67.82	&83.29	&48.88	&58.12	&78.12	\\
				\rule{0pt}{8pt}&1&57.71	&66.94	&83.22	&50.20	&60.18	&78.83	\\
				\rule{0pt}{8pt}&5&\textbf{58.38}	&\textbf{67.23}	&\textbf{83.68} &\textbf{50.80}	&\textbf{60.77}	&\textbf{79.38}	\\
				\rule{0pt}{8pt}&10&58.03  &67.12  &83.29	&50.71 &60.25 &79.23	\\	
                \rule{0pt}{8pt}&50&54.87  &66.47  &82.37	&47.33	&56.76	&75.48	\\
				\hline
			\end{tabular}
	\end{center}
\end{table}
\begin{table}[t]
	\begin{center}
		\caption{Results of PD-Net with different head numbers.}
		\label{tab:issnet_head_num}
			\begin{tabular}{c|ccccccc}
				\hline
				&&\multicolumn{3}{c}{Point Cloud}	&\multicolumn{3}{c}{Image}\\
				\rule{0pt}{8pt}&$H$	&mIoU	&mAcc	&OA	&mIoU	&mAcc	&OA\\
				\hline
				\multirow{3}{*}{20\%}
				\rule{0pt}{8pt}&3&62.08  &70.11  &85.58  &57.70  &67.52  &81.87  \\
				\rule{0pt}{8pt}&4&\textbf{63.38}	&\textbf{71.61}	&\textbf{86.28}	&\textbf{60.17}	&\textbf{70.78}	&\textbf{83.29}	\\	
				\rule{0pt}{8pt}&5&62.16  &70.19  &85.57  &56.44  &65.53  &82.04  \\
				\hline
				\multirow{3}{*}{10\%}
				\rule{0pt}{8pt}&3&56.20	&66.19	&82.46	&49.23	&58.89	&78.75	\\
				\rule{0pt}{8pt}&4&\textbf{58.38}	&\textbf{67.23}	&\textbf{83.68}	&\textbf{50.80}	&\textbf{60.77}	&\textbf{79.38}	\\
				\rule{0pt}{8pt}&5&56.47	&66.43	&82.51	&48.85	&58.19	&78.23	\\	
				\hline
			\end{tabular}
	\end{center}
\end{table}
\begin{table*}[b]
	\setlength{\abovecaptionskip}{0.1cm}
	\setlength{\belowcaptionskip}{-0.1cm}
	\begin{center}
		\caption{Results of PD-Net and Baseline models with different voxel sizes.}
		\label{tab:issnet_voxel_size}
			\begin{tabular}{c|lccccccc}
				\hline
				&&\multicolumn{3}{c}{Point Cloud}	&\multicolumn{3}{c}{Image}\\
				\rule{0pt}{8pt}&Model	&mIoU	&mAcc	&OA	&mIoU	&mAcc	&OA &Time\\
				\hline
				\multirow{4}{*}{20\%}
				\rule{0pt}{8pt}&Baseline (5cm)&59.31	&67.92	&84.13	&45.04	&59.20	&74.96&	1.3s\\
				\rule{0pt}{8pt}&Baseline (2cm)&59.45	&70.35	&83.73	&49.84	&61.00	&76.52&3.1s\\
				\rule{0pt}{8pt}&PD-Net (5cm) &63.38	&71.61	&86.28	&60.17	&70.78	&83.29&	2.5s\\
				\rule{0pt}{8pt}&PD-Net (2cm)&\textbf{64.72}  &\textbf{75.68}  &\textbf{88.17} &\textbf{62.42}  &\textbf{73.54}  &\textbf{86.63}&7.3s \\	
				\hline
				\multirow{4}{*}{10\%}
				\rule{0pt}{8pt}&Baseline (5cm)&52.27	&61.19	&80.81	&43.00	&55.09	&72.75&	1.3s\\
				\rule{0pt}{8pt}&Baseline (2cm)&55.00	&65.16	&81.22	&44.68	&56.53	&72.73&3.1s\\
				\rule{0pt}{8pt}&PD-Net (5cm)&58.38	&67.23	&83.68	&50.80	&60.77	&79.38&2.5s\\
				\rule{0pt}{8pt}&PD-Net (2cm)&\textbf{59.05}  &\textbf{71.01}  &\textbf{85.50}	&\textbf{58.16} &\textbf{68.63} &\textbf{82.80}&7.3s	\\	
				\hline
			\end{tabular}
	\end{center}
\end{table*}
\textbf{Effect of the weight for consistency loss.} As seen in~\eqref{equ:issnet_loss_all}, the loss weight $\lambda_{c}$ affects the balance between the cross-entropy loss terms and $\mathcal{L}_{c}$.
We evaluate the proposed PD-Net with $\lambda_{c}=\{0,0.2,1,5,10,50\}$ to estimate the insensitive range of $\lambda_c$.
The corresponding results are reported in Table~\ref{tab:issnet_loss_weight}, which indicate that our method is relatively insensitive to $\lambda_{c}$ when $\lambda_c$ ranges in $[0.2,10]$ and the performances drop evidently when $\lambda_c$ is set to extreme values (i.e., $\lambda_c=\{0,50\})$.
The model achieves the best performances in most cases when $\lambda_c=5$.

\begin{table}[!t]
\caption{Semantic segmentation results on NYUv2 using dense pixel-level classification accuracy metric.}
\label{nyu_res}
    \centering
\begin{tabular}{lc}\toprule
 \textbf{NYUv2}          & Accuracy \\\midrule
 SceneNet\cite{Handa2015UnderstandingRI} & 52.5\\
 Hermans \textit{et al.} \cite{Hermans2014Dense3S}& 54.3\\
 SemanticFusion\cite{McCormac2016SemanticFusionD3} & 59.2\\
 ScanNet\cite{2017scannet} & 60.7\\
3DMV\cite{20183dmv}        &    71.2      \\
BPNet\cite{2021bpnet}        &    73.5      \\
SemAffiNet \cite{Wang2022SemAffiNetST} & 78.3\\ \midrule
PD-Net (20\%) &    71.7    \\ \bottomrule
\end{tabular}
\end{table}

\textbf{Effect of head number.} As stated in~\cite{2017attention}, multi-head attention allows the model to jointly attend to information from different representation subspaces at different positions, which indicates that more heads could enhance the representation ability of the model.
However, as revealed in~\cite{Michel2019AreSH}, the majority of attention heads can be removed without deviating too much from the original performance and most heads are redundant given the rest of the model at test time. 
And too many heads may result in overfitting considering the strong representation ability on the training set.
Thus, the head number $H$ in the DMF module affects the quality of the fused features and the performance of the model.


Here, we evaluate the proposed PD-Net with $H=\{3,4,5\}$.
And the corresponding results are reported in Table~\ref{tab:issnet_head_num}.
As seen from this table, compared with the results when $H$ is set as 4, the performances of 3D and 2D segmentation degrade when $H$ is set as 3 or 5.
This phenomenon is consistent with the revealed points in~\cite{2017attention} and~\cite{Michel2019AreSH}, which indicates that an appropriate head number needs to be set.

\textbf{Effect of voxel size.} In previous experiments, we set the voxel size to 5cm for efficient training.
We evaluate PD-Net and baseline models with voxel size = \{5cm, 2cm\}, and the corresponding results are reported in Table~\ref{tab:issnet_voxel_size}.
The results show that decreasing voxel size could simultaneously improve the performances
of 3D and 2D segmentation, demonstrating that fine-grained voxels could provide higher-quality 3D information and better boost the 2D semantic segmentation. But in the meanwhile, smaller voxel size inevitably brings higher computational cost and causes longer forward time, as seen in the last column of Table~\ref{tab:issnet_voxel_size}. 

\subsection{PD-Net on NYUv2}
The NYUv2 dataset is a widely-used RGB-D dataset, which contains 1449 densely annotated pairs of aligned RGB and depth images.
Following 3DMV \cite{20183dmv}, BPNet \cite{2021bpnet}, and SemAffiNet \cite{Wang2022SemAffiNetST}, we additionally evaluate PD-Net on NYUv2 dataset \cite{Silberman2012IndoorSA} by converting the depth images to pseudo point clouds according to the camera's pose matrix.
We adopt the 13-class configuration for a fair comparison with the comparative methods \cite{Handa2015UnderstandingRI, Hermans2014Dense3S, McCormac2016SemanticFusionD3, 2017scannet, 20183dmv, 2021bpnet}.

We utilize the pixel-level classification accuracy metric and report the results in Table~\ref{nyu_res}. 
As seen from this table, our proposed PD-Net achieves comparable results with the compared fully-supervised RGB-D and joint 2D-3D methods.
The results on the NYUv2 dataset demonstrate the effectiveness and generality of PD-Net.

\section{Conclusions}
\label{sec: con}
We propose a parallel dual-stream network, called PD-Net, to handle the semi-supervised dual-modal semantic segmentation task.
The proposed PD-Net consists of two parallel streams (i.e., original stream and pseudo-label prediction stream), in which the 3D and 2D encoder-decoder branches are used to extract 3D and 2D features respectively, and multiple dual-modal fusion modules are used to fuse the multi-scale dual-modal latent features.
The pseudo-label optimization module is explored to improve the quality of the pseudo labels output by the pseudo-label prediction stream.
Experimental results demonstrate that the proposed PD-Net not only outperforms the comparative semi-supervised methods but also achieves competitive performances with some fully-supervised methods in most cases.



\bibliographystyle{IEEEtran}
\bibliography{paper.bib}

\newpage

\vfill

\end{document}